\documentclass[letterpaper, 10 pt, journal, twoside]{ieeetran}  

\IEEEoverridecommandlockouts                              

\usepackage{graphics} 
\usepackage{epsfig} 
\usepackage{times} 
\usepackage{amsmath} 
\usepackage{amssymb}  
\usepackage{bm}
\usepackage{orcidlink}
\usepackage{subfloat}

\usepackage{svg}
\svgpath{{./figs/}}

\usepackage{algorithm}
\usepackage{algpseudocode}

\algblock{Input}{EndInput}
\algnotext{EndInput}
\algblock{Output}{EndOutput}
\algnotext{EndOutput}
\algblock{Global}{EndGlobal}
\algnotext{EndGlobal}
\newcommand{\Desc}[2]{\State \makebox[4em][l]{#1}#2}

\usepackage[normalem]{ulem}

\title{
Adaptive and Collaborative Bathymetric Channel-Finding Approach for Multiple Autonomous Marine Vehicles
}

\author{Nikolai Gershfeld $^{*,1}$ \orcidlink{0000-0003-1361-6165}, Tyler M. Paine $^{*,1,2}$ \orcidlink{0000-0001-6071-621X} and Michael R. Benjamin $^{1}$ \orcidlink{0000-0002-2520-6465}
\thanks{Manuscript received: November 5, 2022; Revised March 11, 2023; Accepted May 1, 2023.}
\thanks{This paper was recommended for publication by Editor Pauline Pounds upon evaluation of the Associate Editor and Reviewers' comments.
This work was supported by the United States Military Academy at West Point
and the Office of the Undersecretary of Defense for Research and Engineering (ACC-APG-RTP
W911NF2120206)}
\thanks{$^{*}$These authors contributed equally}
\thanks{$^{1}$Dept. of Mechanical Engineering,
        Massachusetts Institute of Technology, Cambridge, Massachusetts, USA
        {\tt\footnotesize gersni@mit.edu, tpaine@mit.edu, mikerb@mit.edu}}%
\thanks{$^{2}$Applied Ocean Science and Engineering,
        Woods Hole Oceanographic Institution, Woods Hole MA, USA
        }
\thanks{Digital Object Identifier (DOI): see top of this page.}
\vspace{-8mm}
}

\begin{document}

\markboth{IEEE Robotics and Automation Letters. Preprint Version. Accepted May, 2023}
{Gershfeld \MakeLowercase{\textit{et al.}}: Collaborative Channel-Finding for Multiple Autonomous Marine Vehicles} 

\maketitle


\begin{abstract}
This paper reports an investigation into the problem of rapid identification of a channel that crosses a body of water using one or more unmanned surface vehicles (USVs). A new algorithm called Proposal Based Adaptive Channel Search (PBACS) is presented as a potential solution that improves upon current methods. The empirical performance of PBACS is compared to that of lawnmower surveying and Markov decision process (MDP) planning with two state-of-the-art reward functions: Upper Confidence Bound (UCB) and Maximum Value Information (MVI). 
The performance of each method is evaluated through a comparison of the time it takes to identify a continuous channel through an area using one, two, three, or four USVs. 
The performance of each method is compared across ten simulated bathymetry scenarios and one field area, each with different channel layouts.
The results from simulations and field trials indicate that on average multi-vehicle PBACS outperforms lawnmower, UCB, and MVI-based methods, especially when at least three vehicles are used. 

\end{abstract}
\begin{IEEEkeywords}
Multi-Robot Systems, Marine Robotics, Swarm Robotics, Cooperating Robots, Distributed Robot Systems
\end{IEEEkeywords}

\section{INTRODUCTION}

\begin{figure}[ht]
\includegraphics[width=1\columnwidth]{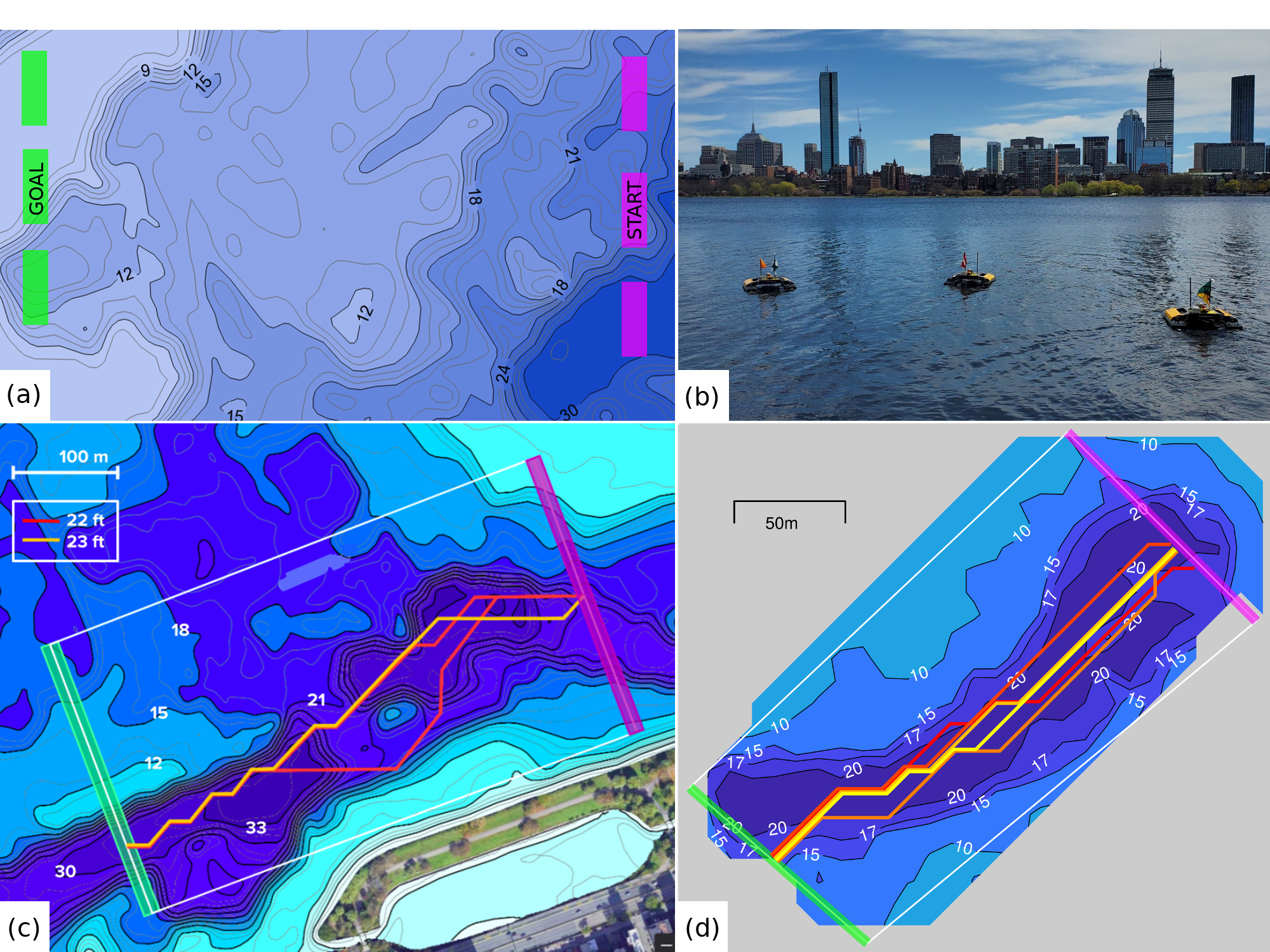}
\vspace{-4mm}
\caption{(a) Illustration of the rapid channel identification problem; quickly find a valid deep channel in an unknown environment that connects any of the start areas (magenta) to any of the goal areas (green). (b) Three of four Heron USVs equipped with low-cost single-beam sonar altimeters used for field testing.  (c-d) Demonstration of our approach to find: (c) a channel deeper than 22ft and 23ft through the deep trench in the Charles River, and (d) a channel deeper than 17ft in Lake Popolopen in New York. The bathymetry map data for (a) and (c) is from \cite{CRAB_Bathy}}
\label{fig:charles_plot}
\vspace{-6mm}
\end{figure}

\IEEEPARstart{A}{utonomous} marine vehicles are important tools for many applications in both civilian and military contexts. One such application is Rapid Environmental Assessment (REA), where a vehicle provides information about the physical environment in an area of interest. This information is then used to inform future missions. In a riverine environment with unknown bathymetry, this can entail quickly identifying a channel that can provide a navigable path through the area as illustrated in Figure \ref{fig:charles_plot}a. We refer to this as the rapid channel identification problem. This paper will focus on the development of a new algorithm to efficiently address this problem. We compare this new method to other state-of-the-art approaches and investigate the utility of using multiple vehicles in the solution.

The simplest way to survey an area is with a lawnmower search pattern. This is the most exhaustive method and will provide a comprehensive overview of the environment. However, for this problem, we are interested in finding a specific feature -- a deep channel -- rather than constructing a complete map. 
To speed up this process, we employ adaptive sampling strategies and robust multi-vehicle task allocation when more than one vehicle is available.

Using adaptive sampling and task allocation methods, we present the Proposal Based Adaptive Channel Search (PBACS) algorithm. 
To the best of the authors’ knowledge, we are the first to demonstrate a decentralized multi-vehicle approach to the channel identification problem, a unique but important challenge in marine engineering.  The PBACS algorithm builds upon a combination of state-of-the-art methods in several areas: non-parametric environmental modeling (Fast Gaussian process regression (GPR) \cite{das}) with decentralized bathymetry map fusion (modified decentralized Kalman consensus (MDKC) \cite{alighanbari}), and multi-objective behavior optimization by interval programming (MOOS-IvP \cite{benjamin})).   The new PBACS algorithm employs an intuition regarding the structure of this particular type of exploration problem with directed search and market-based allocation in a way that outperforms other methods on average.  Our approach is fully decentralized, and we have evidence that the advantages of the method can be realized with just one vehicle or with a cooperative group.

\subsection{Contributions}
The contributions of this research are the following:

\begin{itemize}
    \item The PBACS algorithm, a new specialized method for solving the rapid channel identification problem.
    \item Monte Carlo simulation studies to evaluate the utility of using different amounts of vehicles for PBACS, and to demonstrate better performance than both lawnmower and myopic Markov decision process (MDP) path planning.
    \item Multiple field deployments of the MDP approaches using up to three unmanned surface vehicles (USVs), and deployment of the PBACS approach using up to four USVs. Heron USVs made by Clearpath Robotics are shown in Figure \ref{fig:charles_plot}b, and we report two successful demonstrations in the field where we found channels in the Charles River (Figure \ref{fig:charles_plot}c) and Lake Popolopen in New York (Figure \ref{fig:charles_plot}d).  
\end{itemize}

\section{RELATED WORK} \label{sec:lit_rev}

\subsection{Gaussian process (GP) and Path Planning}

GPR is frequently used in marine adaptive sampling for spatial modeling of the estimated environment, which is then used to inform path planning. Berget et al. \cite{berget} use GP methods to track suspended material plumes. The model is updated continuously throughout the mission, and the path planner uses this estimation to drive the unmanned vehicle to information-rich areas. In Fossum et al. \cite{fossum}, GP modeling is used for adaptive sampling of phytoplankton by modeling the distribution of chlorophyll-a - a common indicator of phytoplankton activity.
Yan et al. \cite{yan} use GPR analysis to guide online path planning for an AUV in an effort to locate hotspots in the field. 
Another work to use GP methods is Stankiewicz et al. \cite{stankiewicz}, where an AUV uses adaptive sampling to explore an area and identify hypoxic zones. The algorithm identifies regions of interest that exhibit some local extrema and concentrates sampling there.

\subsection{Classes of Adaptive Sampling Problems}
Prior work in marine adaptive sampling can be separated into three broad categories: \textbf{1.)} source localization methods as described by Bayat et al. in \cite{bayat}, which include gradient descent \cite{paliotta} and partially observable Markov decision process (POMDP) \cite{flaspohler}, \textbf{2.)} front/boundary determination, which have been demonstrated with single vehicles \cite{zhang}, \cite{petillo15} and multiple vehicles \cite{fiorelli2006multi}, and \textbf{3.)} feature tracking and mapping such as the work of Bennett et al. \cite{bennett}, where a simulated AUV is used to adaptively map bathymetric features like trenches or specific contours. Of these categories, the channel-finding problem explored in this paper is most closely related to the last, but is not specifically addressed in the literature.

\subsection{Multi-Vehicle Considerations}
One of the main considerations in multi-vehicle missions is their formation. When the vehicles all explore the entire field, they can be in fixed formations such as in a leader-follower strategy employed by Khoshrou et al. \cite{khoshrou} and Paliotta et al. \cite{paliotta},  or more flexible approaches such as generating different sailing directions for each vehicle, as proposed by Yan et al. in \cite{yan}. 

When the field is explicitly divided, the divisions can either be predetermined or dynamic. An example of the latter is Kemna et al. \cite{kemna}, where dynamic Voronoi partitioning is used to divide the field among a group of AUVs.  In general, flexible formations and dynamic divisions are more robust to single vehicle failures, which is part of the motivation for our approach. 

Information sharing is a critical component of multi-vehicle systems, and communications can be range restricted or otherwise time-varying, especially in the marine domain.
To address this problem, we use consensus protocols and algorithms for multi-vehicle coordination developed by Ren et al. \cite{ren2} and Alighanbari et al. \cite{alighanbari}.  In particular, we use the robustness properties of the MDKC  developed by Alighanbari et al. \cite{alighanbari}, which removes potential biases that occur when the agents in the network are not fully connected.

In many cases, vehicles need to coordinate new tasks/roles that arise as the mission progresses. One class of algorithms for solving this problem of task allocation is auction-based algorithms, also referred to as market-based algorithms. 
The implementations of these algorithms can rely upon a centralized repository, as described by Bertsekas \cite{bertsekas}. They can also be distributed, as described by Michael et al. \cite{michael} and Zavlanos et al. \cite{zavlanos}, which can be augmented with a consensus protocol to resolve conflicts \cite{brunet2008consensus,raja2021communication}.

\subsection{Multi-robot systems (MRS) for other applications}
Recently, solutions to the general problem of collaborative exploration of an entire 3D map have been demonstrated with aerial vehicles \cite{Tan2023RAL} and a mix of aerial and ground vehicles \cite{CERBERUS20222}.  
However, our decentralized approach is designed to concentrate search in areas where a channel may still be viable. 
Furthermore, although the CERBERUS system in \cite{CERBERUS20222} was the most successful in the DARPA Subterranean Challenge, the system used a centralized map server as the arbiter of information transmitted among the individual robots.   
Other decentralized approaches such as those reported in \cite{Hou2022RAL} and \cite{Toumieh2022RAL} use low-level control, which requires accurate models of the dynamics of each agent or conservative estimates that reduce optimality.  
In the case of marine vehicles, these models are more complex, and they operate in environments that are stochastic due to wind, waves, and currents \cite{fossen2011handbook}. 
Due to these limitations, we instead implement multi-objective optimization via MOOS-IvP and control at the individual level.
Finally, we emphasize our repeated experimental demonstrations of  a multi-robot system in the field; the vehicles perform decentralized estimation and planning with limited computing power, low-cost single-beam sonars, and communication limitations.

\section{ENVIRONMENTAL MODELING} \label{sec:meth}

We represent the bathymetry data in a discrete set of $m$ grid-cells. We define the vectors $\vec{z} \in \rm I\!R^m$ and $\vec{\sigma} \in \rm I\!R_+^m$ as the depth and the variance (respectively) of each cell. The location of the center of the $i^{th}$ cell is denoted as $\vec{x}_i\in \rm I\!R^2$.

\subsection{Single-Beam Depth Sensor}
We assume the depth directly below each vehicle can be measured by a single-beam acoustic depth sensor. For field work, we used the Ping Sonar Altimeter and Echosounder made by Blue Robotics for its low cost and commercial availability. A simulated sensor with comparable accuracy and noise was used for simulation studies. 

The sensor was mounted onto the bottom plate of each Heron USV at the lowest possible depth that it remains submerged at all times without contacting the dock. The sensor was mounted directly below the GPS sensor in order to ensure that the distance readings are associated with the correct position. The sensor was then connected to the Heron's payload autonomy computer and integrated into the MOOS system described in Section~\ref{section:implement_dets}. We set the sampling rate of the sensor to 10 Hz.

\subsection{Efficient GPR}

We use GPR on each vehicle to calculate an estimate for the bathymetry field within the grid from the collection of local measurements. The GPR implementation used in this system is adapted from the "Fast GPR" developed by Das et al. \cite{das}. The Fast GPR aims to speed up computation by developing estimators on subsets of the dataset. Using the standard implementation of GPR on a dataset of size $N$ yields $O(N^3)$ time complexity. By choosing $k$ subsets of size $N_s < N$, the time complexity of Fast GPR is reduced to $O(kN_s^3)$. One limitation of this method is that the variance becomes skewed, but  we address this minor complication in Section \ref{sec:path_exp} with the threshold in (\ref{eq:threshvar}). 

We use the Gaussian kernel
\begin{equation} \label{eq:rbf}
    \mathcal{K}(\vec{x}_i, \vec{x}_j) = exp \Big( - \frac{d(\vec{x}_i,\vec{x}_j)}{2 l^2} \Big),
\end{equation}
where $d(\vec{x}_i,\vec{x}_j)$ is the Euclidean distance. We determined experimentally that $l \approx 28.8ft$ works well in field testing. 

\subsection{Decentralized Map Fusion}\label{sec:kalman_cons}

For multi-vehicle coordination, we use the Modified Decentralized Kalman Consensus (MDKC) reported by Alighanbari and How \cite{alighanbari}. Although we extensively test our approach with USVs that can more easily communicate, our approach is intended to be used in multi-vehicle systems that include unmanned underwater vehicles (UUVs), which have limited communications while submerged.  For that reason, we use a periodic consensus to combine estimates of the map from each vehicle in a way that requires limited communication bandwidth, works without a fully connected communications graph, and is robust to intermittent communication failures.

A consensus can be reached even if the group does not form a fully connected graph, which happened periodically during field operations as vehicles temporarily drop out of communication. 
\vspace{4mm}

For $n$ agents $\mathcal{A} = \{\mathcal{A}_i,...,\mathcal{A}_n\}$, the solution for an agent $\mathcal{A}_i$ at time $t+1$ is given by:
\begin{align}
\bm{P_i}(t+1) =& \{[\bm{P_i}(t)+\bm{Q}(t)]^{-1} \nonumber \\
&+ \sum_{j=1}^{n}(g_{ij}(t)[\mu_j(t)\bm{P_j}(t) ]^{-1})\}^{-1}
\end{align}
\begin{align}
\vec{z}_i(t+1) &= \vec{z}_i(t) \nonumber\\
& + \bm{P_i}(t+1)\sum_{j=1}^{n}\{g_{ij}(t) \cdot [\mu_j(t)\bm{P_j(t)} ]^{-1} \nonumber \\
& \cdot (\vec{z}_j(t)-\vec{z}_i(t))\} 
\end{align}
\begin{equation}
\mu_j(t) = \sum_{k=1, k \neq j}^{n}g_{kj}(t)
\end{equation}
where $\bm{P_i}$ is the covariance matrix assembled using the radial basis function kernel (\ref{eq:rbf}) and the grid variance $\vec{\sigma}$, $\bm{Q}(t)$ is the process noise (used only when a vehicle becomes completely disconnected and must complete the consensus on their own), $\vec{z}_i$ is the agent's own information, $g_{ij}$ is the adjacency matrix of the communication graph between agents $\mathcal{A}_i$ and $\mathcal{A}_j$, and $\mu_j(t)$ is a scaling factor associated with agent $\mathcal{A}_j$. 

\section{CHANNEL SEARCH} \label{sec:PBACS}
\begin{figure}[h]
\centering
\includegraphics[width=\columnwidth]{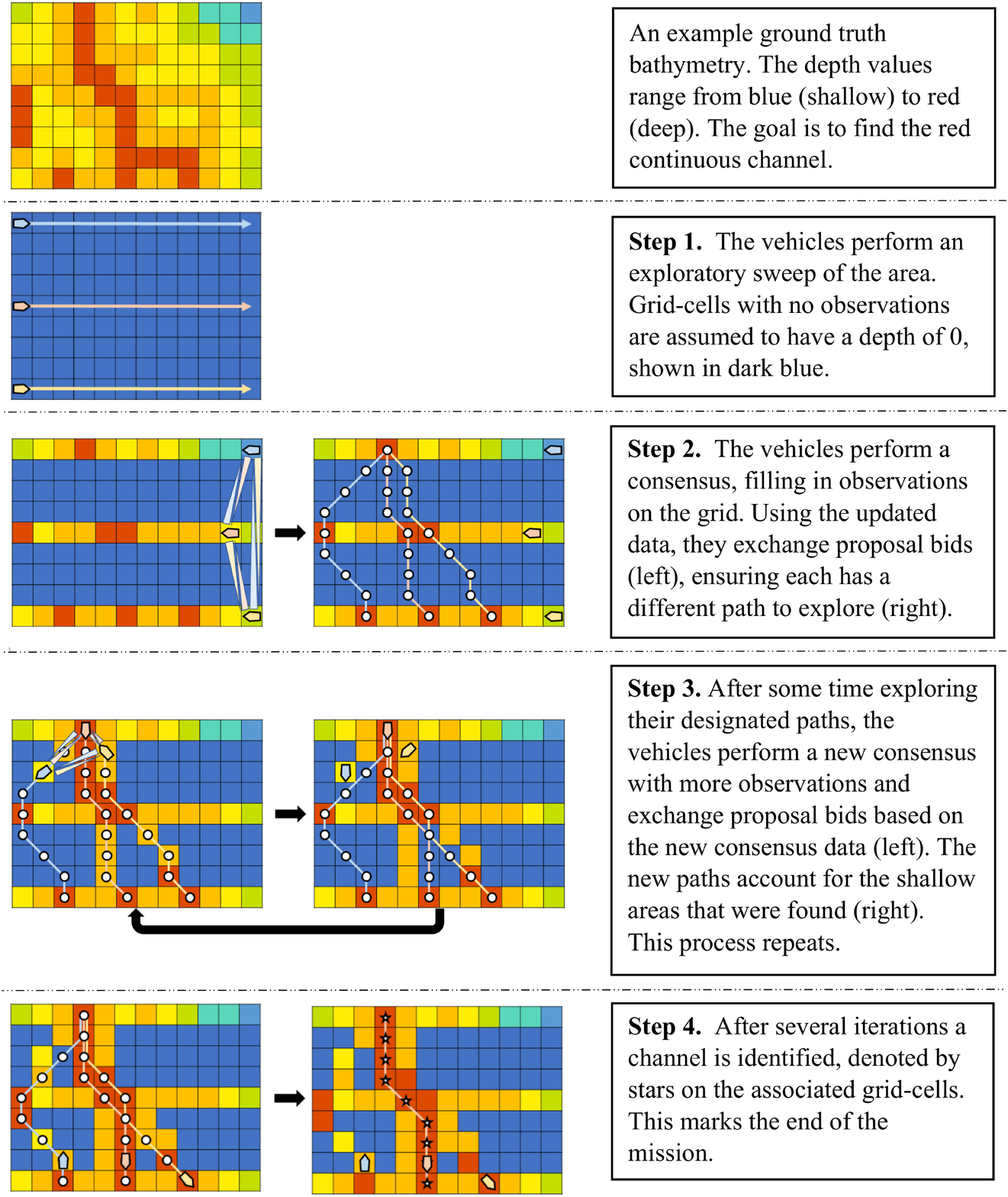}
\vspace{-4mm}
\caption{The simplified steps of PBACS.}
\label{fig:hybrid_process}
\vspace{-4mm}
\end{figure}

Here we present our main algorithm, a more specialized method for solving the channel identification problem.
There are two stages to this approach: an exploratory sweep and a search along candidate paths. A simplified version of the process is illustrated in Figure~\ref{fig:hybrid_process}, and the proposal algorithm is reproduced in Algorithm \ref{algo::PBACS}. 

\subsection{First Stage - Exploratory Sweep}
The first stage of PBACS is an exploratory sweep of the area. Other adaptive sampling works have also employed an initial exploratory lawnmower sweep to seed the remainder of the search \cite{bennett}, \cite{paliotta}. For our problem, we can use such a sweep to help save time by eliminating areas that do not meet the necessary depth criteria, thereby directing the search toward more likely channel regions (Step 1 in Figure~\ref{fig:hybrid_process}). For a single-vehicle mission we set the initial sweep to cover the start area of the grid, and for a two-vehicle mission we cover the start and end goal areas. The sweep area for each subsequent vehicle is an equally spaced line in between. 

\subsection{Second Stage - Path Exploration} \label{sec:path_exp}
The second stage after the sweep is the path exploration stage, shown in Algorithm \ref{algo::PBACS}. Vehicles enter this stage after they complete their initial sweep, and this transition often occurs asynchronously due to the difference in transit times to the initial sweep locations. 

The goal of this stage is to identify and explore candidate paths that may still be viable. Each vehicle proposes a path between the start and goal regions that may be part of a viable channel (Step 2 in Figure~\ref{fig:hybrid_process}). These candidate paths can be generated using any path planner; here we used a common variant of the A* planner which searches for a path that connects any of  several start and goal locations.

The depth along a candidate path must not be shallow, and we capture this criterion in our search problem with simulated obstacles. Grid-cells are considered obstacles only if they have a sufficiently low variance \emph{and} are too shallow. This way, candidate paths are made up of cells that are either certainly deep enough, or that we are uncertain about how deep they actually are.
The variance threshold for considering a shallow grid-cell to be an obstacle is calculated dynamically as a percentage of the range between the current minimum and maximum variance across the grid, i.e. 
\begin{equation} \label{eq:threshvar}
    \sigma_{th} = \sigma_{min} + \eta (\sigma_{max} - \sigma_{min}).
\end{equation}
where the probability distribution is computed using the MDKC algorithm as described in Section \ref{sec:kalman_cons}.
This threshold must be set high enough to exclude shallow grid-cells with low variance, typically those that have been directly measured by at least one vehicle. The threshold should also be low enough to not exclude shallow grid-cells with high variance, typically those that were interpolated. We experimentally determined $\eta = 0.33$ to be sufficient for both simulation and fieldwork.

Upon completion of each GPR estimate and consensus, the path is rechecked to ensure that no new obstacles were found on it and that it is still optimal (Steps 2 and 3 in Figure~\ref{fig:hybrid_process}). If a more optimal and obstacle-free path is found, the vehicle switches to this path (Step 3 in  Figure~\ref{fig:hybrid_process}). We periodically check for the existence of a valid continuous channel, which marks the end of the mission (Step 4 in  Figure~\ref{fig:hybrid_process}).

\begin{subfigures}
\begin{figure} 
\vspace{2mm}  
\end{figure}
\begin{algorithm}[t]
\caption{Pseudocode of the main PBACS algorithm}
\label{algo::PBACS}
\begin{algorithmic}
\Global \textbf{ variables}
\Desc{$\mathbf{p}_{curr}$}{\parbox[t]{.7\columnwidth}{vector of waypoints of current path $[p_1,p_2,\hdots,p_N]$}}
\vspace{1mm}
\Desc{$\mathbf{N}_{\mathbf{p}}$}{number of waypoints/vertices in path $\mathbf{p}$}
\Desc{$\mathbf{P}$}{\parbox[t]{.7\columnwidth}{vector of tuples $(\mathbf{p}_i,c_i,t_i)$ containing received proposals, costs, and time received}}
\EndGlobal
\end{algorithmic}

\begin{algorithmic}[1]

\Procedure {PBACS}{}
\State initialize $won \leftarrow false$ 
\Repeat
\State \parbox[t]{.8\columnwidth}{Handle incoming messages, proposals, \textcolor{black} {checking for new consensus map data from GPR and MDKC processes}}
\vspace{1mm}
\If {new consensus data}
\State \textcolor{black}{Build search grid from $\vec{z}(t)$, $\bm{P_i}(t)$, $\sigma_{th}$ (\ref{eq:threshvar})}
\State $chnl\_fnd \leftarrow$ check for channel \Comment{A*}
\State $waited \leftarrow time\_since\_last\_prop > t_{wait}$
\If{$\neg chnl\_fnd$ \textbf{and} $\neg won$ \textbf{and} $waited$}
\State $won = \textsc{checkProposals}(\mathbf{P})$

\If {$\neg won$}
\State $\mathbf{p}_{new} \leftarrow $ find candidate path \Comment{A*}

\If {$\mathbf{p}_{new} \not= [~]$}
\State $cond1 = \mathbf{p}_{curr}$~has new obstacles
\State $cond2 = N_{\mathbf{p}_{curr}} > N_{\mathbf{p}_{new}}$
\If{$cond1$~\textbf{or}~$cond2$}
\State $\mathbf{p}_{prop} \leftarrow \mathbf{p}_{new}$
\Else
\State $\mathbf{p}_{prop} \leftarrow \mathbf{p}_{curr}$
\EndIf
\Else
\State \textbf{exit}, use MDP planning
\EndIf

\State $c_{prop} \leftarrow$ min distance to $\mathbf{p}_{prop}$
\State send $\mathbf{p}_{prop}$ and ${c}_{prop}$ to other vehicles

\ElsIf{$won~\textbf{and}~ \mathbf{p}_{prop} \not= \mathbf{p}_{curr}$}
\State $p_a \leftarrow$ get closest waypoint in $\mathbf{p}_{prop}$
\State $\mathbf{p}_{curr} = \textsc{getDirection} (\mathbf{p}_{prop},p_a)$
\State publish $\mathbf{p}_{curr}$ to waypoint behavior
\EndIf

\EndIf
\EndIf

\Until{$channel\_found$}
\EndProcedure
\end{algorithmic}
\end{algorithm}
\end{subfigures}

\subsection{Path Search Direction}
\begin{figure}[ht]
\centering
\vspace{4mm}
\includegraphics[width=0.55\columnwidth]{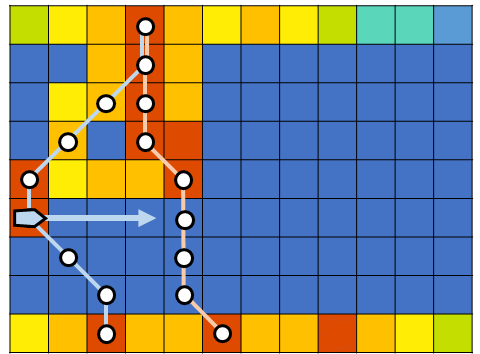}
\caption{An example of a vehicle switching to a new candidate path, in a mission where it aims to identify a continuous deep channel. The desired channel depth is shown in red, and areas with no samples and high variance are shown in blue. The vehicle has encountered an obstacle on $\mathbf{p}_{curr}$, the path it was exploring (left path), and has found $\mathbf{p}_{prop}$, a new candidate path (right path). The blue arrow points in the direction of $p_a$, the vehicle's first waypoint on the new path. There are two choices of directions in which to traverse the path, $\mathbf{p}_{r}$ (reverse/up) or $\mathbf{p}_{f}$ (forward/down). Since $\mathbf{p}_{f}$ would pass through more high variance grid-cells, this direction is chosen. }
\label{fig:direction}
\vspace{-6mm}
\end{figure}

Once a vehicle has a new candidate path to explore, its first waypoint on that path is the point with the shortest Euclidean distance to its current position. Since this is most likely not an endpoint of the candidate path, we must choose the direction in which the vehicle will traverse the path. 
The first condition we check is whether we have an unexplored endpoint. Due to the preliminary sweep stage, this condition would only happen in the single-vehicle case. 
The second condition is checking which side of the path has a higher variance, so that the vehicle is directed toward less explored regions. This is done by comparing the sum of the variances associated with either direction, with an exponentially increasing discount factor applied to points that are further away. The direction with the higher reward is chosen. The vehicle traverses the waypoints in this direction of the path until it reaches the endpoint. If no obstacles are found, the vehicle reverses to cover the other direction. An example of direction choosing is illustrated in Figure~\ref{fig:direction}.

\subsection{Proposal Bidding}
In the multi-vehicle case, vehicles allocate viable paths among themselves by sharing their locally found path with other vehicles as a proposal with an associated cost, and then checking for conflicts. 
When vehicles update their consensus estimate of the field, each vehicle finds the most optimal potentially viable path. The vehicle then proposes this candidate path to the other vehicles by placing a proposed bid with the cost of transiting to the path. For the cost function, we use the Euclidean distance to the closest point on the path. Each vehicle compares its own proposal to the ones it receives from the others. If there are no conflicts, the vehicle transits to this path and uses a waypoint following behavior to survey along the line. If there is a conflict, the vehicle with a lower-cost bid wins. The losing vehicle must propose a new path and go through another iteration of bidding with all other vehicles that remain unassigned. The new proposal for subsequent iteration rounds takes into account the assigned paths by considering the cells in those paths to be obstacles, ensuring that new proposals do not overlap with paths assigned in the previous iteration. In the case where there are more vehicles than there are potential paths, a vehicle with no path will revert to using an MDP-based surveying mode to explore more of the field until a path becomes available. The proposal process is repeated upon each completion of a consensus.

\section{EXISTING METHODS FOR COMPARISON}
\subsection{Lawnmower Survey}
The most common method of surveying a region is with a lawnmower pattern, also referred to as a boustrophedon in coverage path planning \cite{choset}. 
However, the success of lawnmower surveys in quickly identifying a channel is highly dependent on the (lucky) choice of starting locations and distributions of vehicles that fit the underlying map. In this work, we are interested in approaches that do not rely on luck and perform better in aggregate. 

\subsection{Markov decision processes}
\label{section:mdp}
We can formulate the problem as an MDP \cite{bellman}. 
An MDP consists of the following: a set of states $S$ with an initial state $s_0$, a set of actions in each state $A$, a transition model $P(s'|s,a)$ giving the probability that an action $a$ in state $s$ will lead to state $s'$, and a reward function $R(s)$.

A solution to the MDP is in the form of a policy $\pi$, where $\pi(s)$ gives the recommended action at state $s$. The optimal policy $\pi^*(s)$ for a state is given by:
\begin{equation}
\pi^*(s) = \underset{{a \in A(s)}}{\mathrm{argmax}}\, \sum_{s'} P(s'|s,a) U(s')
\end{equation}
where $U(s)$ is a utility function. This utility function can be estimated using value iteration, which is computed by iteratively applying a Bellman update.

For our implementation, we define the root state to be the grid-cell corresponding to the vehicle's current position and heading. For the actions, to account for the underactuated vehicle's maneuverability we define three possible next grid-cells based on the vehicle's heading: ahead, ahead-left, or ahead-right. 
The state transition model $P(s'|s,a)$ captures the uncertainty in the stochastic environment with wind and waves, and the specific details are reported in \cite{gershfeld}.

An MDP can be implemented with different reward functions. We compare the performance of two different functions, UCB and MVI.
UCB as a reward criterion is a common, state-of-the-art choice in online decision-making \cite{srinivas}. 
An alternative to UCB reward calculation is MVI, defined by Flaspohler et al. \cite{flaspohler}, based on the Max-value Entropy Search (MES) criterion defined by Wang et al. \cite{wang}. As shown by Flaspohler et al., over time the MVI reward converges to the global maximum, while UCB rewards high-value regions more uniformly. 

\section{IMPLEMENTATION DETAILS} \label{section:implement_dets}
We implemented our decentralized autonomy with MOOS-IvP \cite{benjamin}, an open source C++ robotic autonomy software. 
The Mission Oriented Operating Suite (MOOS) is a robotic middleware that uses a publish-subscribe architecture for communication between separate processes (MOOS Apps). 
One of these processes is the Interval Programming (IvP) Helm, which provides behavior-based autonomy in the frontseat-backseat paradigm.

A consensus manager MOOS App was created to manage requests, iterations, and timeouts. An instance of the manager runs on every vehicle to achieve fully decentralized estimation. Due to the computational limits of the on-board computer, another C++ class was written to handle asynchronous sensor data and to efficiently distribute the computation for the Fast GPR over several process cycles. For instance, on each round of sampling we computed the inverse of the covariance matrix via the Cholesky decomposition and stored it for use in subsequent process cycles.

Due to computational constraints of fieldwork hardware, we implemented the MDP as a myopic planner with a limited look-ahead depth. Based on this information, the vehicle determines the current optimal path to take. The utility function is recalculated once it reaches the next grid-cell.

\section{SIMULATED EXPERIMENTS} \label{sec:results}
\begin{figure}[h]
\includesvg[width=\columnwidth]{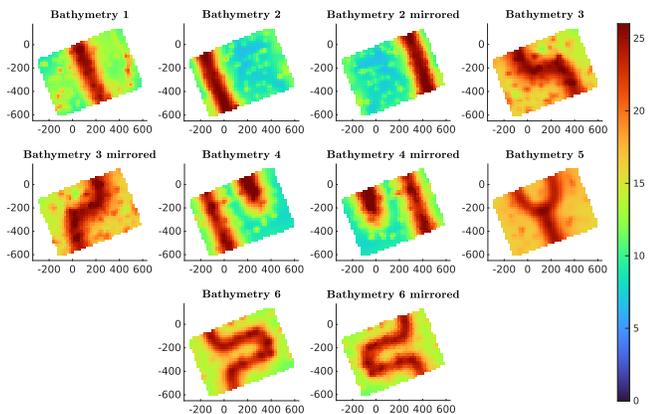}
\vspace{-4mm}
\caption{The ten bathymetry grid scenarios for simulation testing, where the start and goal regions are along the upper and lower edges, respectively. The x and y axis markers represent distances in meters.}
\label{fig:sim_gt}
\end{figure}

To test algorithm performance in simulation, we generated six possible bathymetry scenario maps. For non-symmetric bathymetry configurations, we also used mirrored versions to check for biases that different orientations can introduce. Including these mirrored versions, we tested a total of ten scenarios, which are shown in Figure~\ref{fig:sim_gt}. Each area represents a 500 meter by 750 meter rectangle with 20 meter by 20 meter grid-cells. The grids are tilted to represent the path planning algorithms not being dependent on a perfectly horizontal or vertical grid, to account for possible shoreline bounds. The depth values over all scenarios range from 6 feet to 26 feet.

We conducted a total of 640 Monte Carlo simulation trials at a depth threshold of 20 feet, using one to four vehicles. These trials were spread as follows: for each bathymetry scenario, we ran one lawnmower mission per vehicle number and five missions each of PBACS, UCB, and MVI per vehicle number. Each vehicle's speed is set to 2.4 m/s. For the two MDP approaches, we use a look-ahead depth of 6 cells, recalculated every time a vehicle enters a cell. Each mission ends either when a channel path is found or when the time exceeds 8000 seconds, the length of time for a single vehicle to complete a full lawnmower over the field. 

\subsection{Simulation Results}
We only provide a general summary of performance due to page length constraints; detailed results from each scenario can be found in \cite{gershfeld}. All four approaches consistently found similar channel paths for all scenarios, with the exception of timeouts for all single-vehicle MDP runs and intermittent timeouts for two to four-vehicle MDP and single-vehicle PBACS runs (Table \ref{table:timeout}). Since all paths found are similar in their shape and location, we can compare the success of the missions based on the total time to find the channel and on how much of this time was spent exploring the channel.

\begin{figure}[h]
\centering
\includesvg[width=\columnwidth]{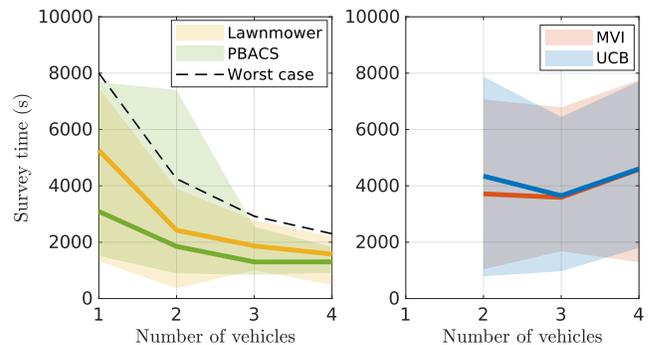}
\vspace{-4mm}
\caption{Means and ranges of mission times across all 10 simulated bathymetry scenarios for the lawnmower, PBACS, and MDP approaches. The timed-out missions listed in Table~\ref{table:timeout} are not represented in the figure. The dashed black line shows the duration of a theoretical mission where the lawnmower covers the entire field. This is not linear due to the field not being evenly divisible by all of the vehicle amounts, and to account for consensus times. PBACS has the fastest mission time on average.}
\label{fig:times_allsims}
\vspace{-2mm}
\end{figure}

On average, the PBACS algorithm identifies the channel faster than the other methods for any number of vehicles from one to four. As shown in Figure~\ref{fig:times_allsims}, the PBACS mission durations are shorter than the lawnmower mission durations on average, but there are some outlier cases of long PBACS missions. The single vehicle case of PBACS times out or is close to timing out for the very curved channel in the Bathymetry 6 scenario and for one orientation of the ``dead end'' channel in the Bathymetry 4 scenario. The two-vehicle case of PBACS also has a long mission duration for the Bathymetry 6 scenario. The three and four-vehicle cases of PBACS consistently perform better than the three and four-vehicle lawnmower surveys, except when a vehicle's lawnmower start location aligns with a straight channel. The lawnmower is fastest when this alignment occurs, but since this cannot be predicted ahead of time on a map with an unknown bathymetry, the lawnmower method cannot reliably provide this result. On the other hand, PBACS maintains the same performance for a channel shape regardless of orientation. The MDP-based methods take the most time to find the channel on average and have much higher variability than the other methods. We can attribute this in part to the limited depth of search that is possible with on-board microprocessors.

\begin{table}[h!]
\begin{center}
\caption{Fraction of simulated missions that timed out}
\label{table:timeout}
\vspace{-2mm}
\begin{tabular}{c c c c c} 
 \hline
\rule{0pt}{2ex}Vehicles & Lawnmower & UCB & MVI & PBACS \\ [0.5ex] 
 \hline 
 \rule{0pt}{2ex}1 & 0/10 & 50/50 & 50/50 & 13/50 \\ 
 2 & 0/10 & 12/50 & 5/50 & 0/50 \\
 3 & 0/10 & 5/50 & 3/50 & 0/50 \\
 4 & 0/10 & 6/50 & 6/50 & 0/50 \\ 
\end{tabular}
\end{center}
\footnotesize 
The MDP approaches consistently had the highest fraction of mission time-outs. Though single-vehicle PBACS missions had some time-outs, this was reduced to zero in multi-vehicle runs.  
\par
\bigskip
\vspace{-4mm}
\end{table}

\begin{figure}[ht]
\centering
\includesvg[width=\columnwidth]{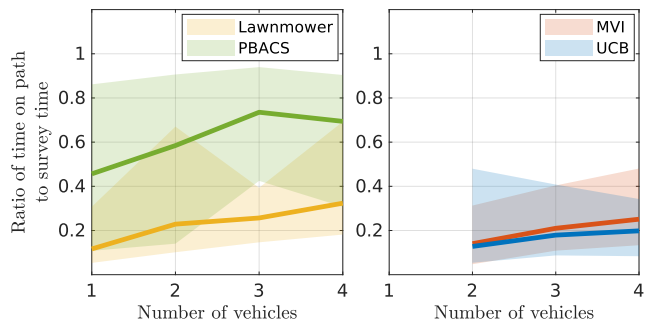}
\vspace{-4mm}
\caption{Ratio of time the vehicles spent on the final channel path to the total survey time across all 10 simulated bathymetry scenarios. The timed-out missions listed in Table~\ref{table:timeout} are not represented in the figure. PBACS missions generally have higher ratios than other approaches, as the vehicles are able to concentrate more quickly on the channel region.}
\label{fig:percents_allsims}
\vspace{-2mm}
\end{figure}

Another measure of mission success is a high percentage of time spent on the final reported channel path. This means that vehicles use the mission time more efficiently on the area of the channel rather than exploring the entire field. In general, more measurements in the channel area increases the accuracy of the end result, given that surface conditions may skew some measurements. As shown in Figure~\ref{fig:percents_allsims}, the two MDP-based approaches consistently have lower percentages than the other two approaches, as the vehicles explore most of the field before completing the mission. The lawnmower missions are generally also low in percentage, with high outliers when a vehicle's start location aligns with a straight channel. The PBACS missions have consistently higher percentages than the rest of the approaches, with some low outliers including the very curved channel in the Bathymetry 6 scenario.

Overall, the results shown here indicate that for these types of environments the optimal amount of vehicles is three when using PBACS. Using three vehicles avoids the possible long mission durations of the one and two-vehicle cases. Adding a fourth vehicle does not offer much improvement in mission duration, and could potentially slightly increase it as well. However, adding a fourth vehicle does reduce variability. These numbers are based on a maximum of one bend in the channel (scenario 6). If more bends were to be added, it is possible that four or more vehicles would be optimal instead.

\section{FIELD EXPERIMENTS}
There were two sets of fieldwork trials. We first performed a comparison study of the four methods with 28 trial runs in a section of the Charles River adjacent to the MIT Sailing Pavilion in spring 2022.
The quantifiable results of these experiments are provided in Section \ref{sec:field_results}.
For the comparison study, the survey area is a 170 meter by 260 meter rectangular grid, with 10 meter by 10 meter grid-cells. This area, chosen for its proximity to the lab space, does not have a clearly defined channel. However, it contains some straight paths down through deep areas as well as some curved paths.

We also repeatedly demonstrated the performance of the PBACS approach to quickly identify a well-defined channel in two bodies of water: the Charles River on the side opposite the MIT Sailing Pavilion in summer 2022 using three vehicles, and the remote Lake Popolopen in New York in fall 2022 using four vehicles. The results of these eight successful demonstrations are shown in Figure \ref{fig:charles_plot}c and Figure \ref{fig:charles_plot}d, respectively. 

\subsection{Field Results} \label{sec:field_results}
As in the simulations, the results for the lawnmower survey are taken from one run. The results of the PBACS algorithm are taken from four runs of each vehicle number. We only have data for the two and three-vehicle cases of both MDP approaches; these results are taken from two runs each. In all missions, the final channel paths found were equally viable. We compare the success of the different missions based on the amount of time to identify a channel, and on the channel path shape.

\begin{figure}[ht]
\vspace{-4mm}
\centering
\includesvg[width=\columnwidth]{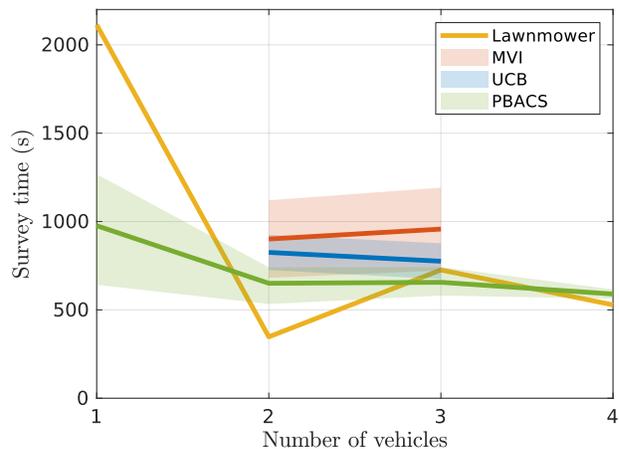}
\vspace{-4mm}
\caption{Field mission time ranges for the MDP and PBACS at the Charles River location. The means are marked with lines.}
\label{fig:ranges_field}
\end{figure}

\begin{figure}[ht]
\centering
\includegraphics[width=1\columnwidth]{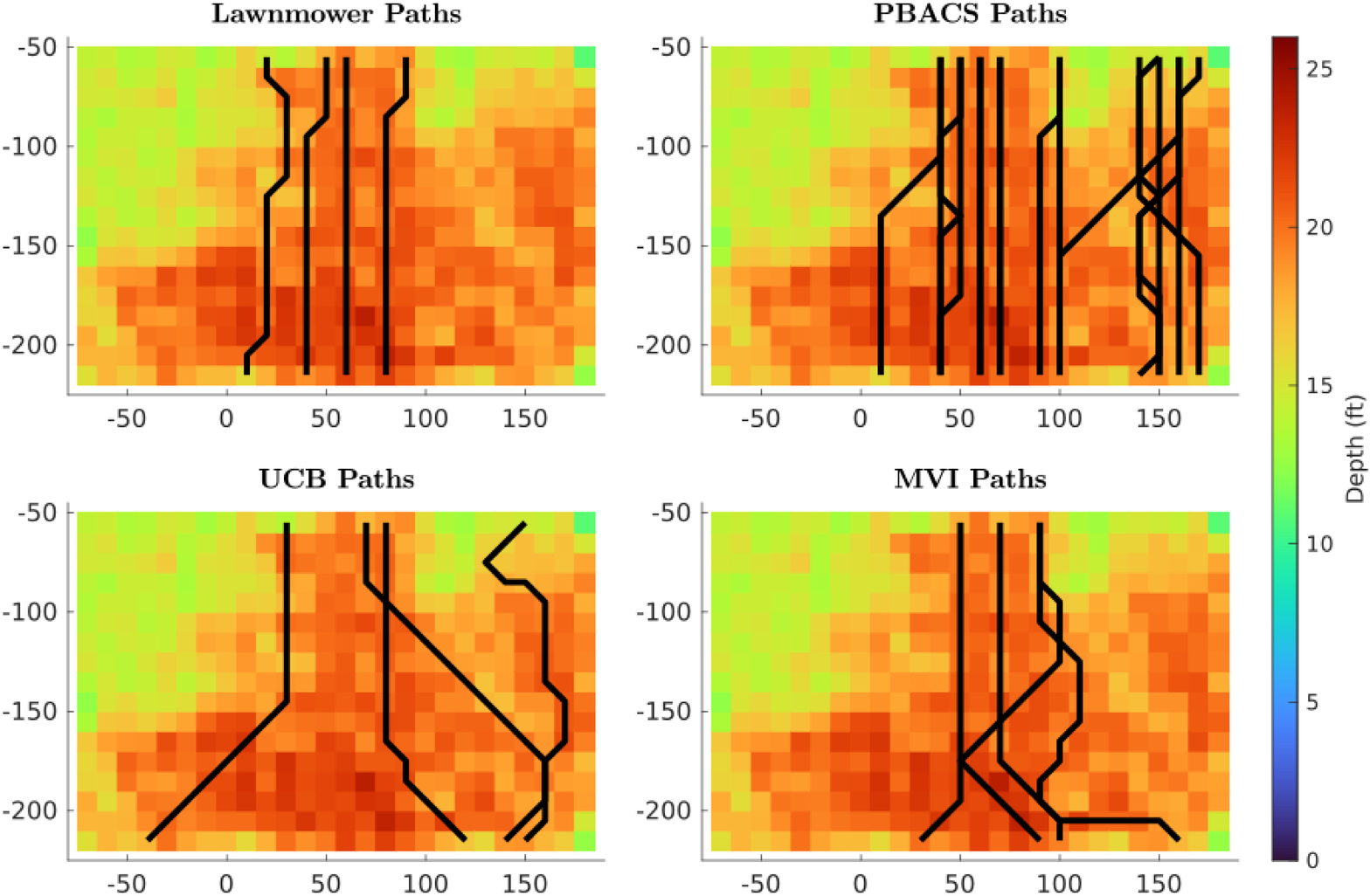}
\vspace{-4mm}
\caption{All final paths found during the Charles River field missions, separated by path planning approach. The x and y axis markers represent distances in meters.  The ground truth grid was averaged from three complete lawnmower surveys. There is still some variability, $\pm 1$ foot, in the water level and surface conditions of each field mission, which accounts for some of the paths passing through grid-cells that appear shallow on the plotted grid.}
\label{fig:paths_field}
\end{figure}

As shown in Figure~\ref{fig:ranges_field}, the PBACS mission duration decreases when the number of vehicles is increased, as does the variation in the mission time. As was found in simulations, when a relatively straight channel path exists, the largest improvement in time comes from increasing the vehicle amount from one to two. For this bathymetry, the channel area aligned particularly well with the two-vehicle lawnmower spacing, so a channel path was identified relatively quickly. As mentioned in the simulation results, the lawnmower method performs best when the spacing works out this way by chance.

As in the simulations, the channel paths found in all field runs were equally viable. However, the field channel paths differed from the simulation channel paths in their shape. This is due to the lack of a clearly defined channel in the field area as opposed to the simulated scenarios. As shown in Figure~\ref{fig:paths_field}, the lawnmower survey and PBACS both have generally straighter paths than the two MDP approaches, since the lawnmower survey runs parallel to this direction and since PBACS checks for the shortest straight paths first. In this way, PBACS outperforms the MDP methods by providing shorter, more efficient crossings if they exist.

\section{CONCLUSION} \label{sec:conclusion}
To the best of our knowledge, this paper describes the first formal investigation into the rapid channel identification problem. This problem is well suited to the use of adaptive sampling. However, our results suggest that it requires fundamentally different approaches than those used in other adaptive sampling problems. Through our simulation and field testing with USVs and single beam altimeters, we found that the PBACS algorithm on average outperforms the lawnmower and both myopic MDP reward functions in multi-vehicle cases.

\addtolength{\textheight}{-5cm}   





\bibliographystyle{IEEEtran}
\bibliography{IEEEabrv, bib/IEEEexample.bib, bib/gersni_thesis.bib}

\end{document}